\def\year{2022}\relax
\documentclass[letterpaper]{article} 
\usepackage{aaai22}  
\usepackage{times}  
\usepackage{helvet}  
\usepackage{courier}  
\usepackage[hyphens]{url}  
\usepackage{graphicx} 
\usepackage{natbib}  
\usepackage{caption} 
\usepackage{comment} 
\DeclareCaptionStyle{ruled}{labelfont=normalfont,labelsep=colon,strut=off} 
\usepackage{algorithm}
\usepackage{algorithmic}

\usepackage{newfloat}
\usepackage{listings}
\floatstyle{ruled}
\newfloat{listing}{tb}{lst}{}
\floatname{listing}{Listing}
\title{Clustering-based Tile Embedding (CTE): A Representation for Level Designs with Skewed Tile Distributions}
 \author{
     Mrunal Jadhav, Matthew Guzdial
 }
 \affiliations{
     Computing Science Department, Amii\\
     University of Alberta\\
     {mrunalsu,guzdial}@ualberta.ca
 }

\begin{document}
\maketitle
\begin{abstract}
There has been significant research interest in Procedural Level Generation via Machine Learning (PLGML), applying ML techniques to automated level generation.
One recent trend is in the direction of learning representations for level design via embeddings, such as tile embeddings. 
Tile Embeddings are continuous vector representations of game levels unifying their visual, contextual and behavioural information.
However, the original tile embedding struggled to generate levels with skewed tile distributions. 
For instance, Super Mario Bros. (SMB) wherein a majority of tiles represent the background. 
To remedy this, we present a modified tile embedding representation referred to as Clustering-based Tile Embedding (CTE).
Further, we employ clustering to discretize the continuous CTE representation and present a novel two-step level generation to leverage both these representations.
We evaluate the performance of our approach in generating levels for seen and unseen games with skewed tile distributions and outperform the original tile embeddings.

\end{abstract}

\section{Introduction}
Procedural Content Generation via Machine Learning (PCGML) involves training machine learning models on existing game data to generate new content such as levels, characters, stories, and music \cite{summerville2018procedural}. 
Due to limited publicly available datasets, particular games have received a disproportionate amount of attentions from PCGML researchers, especially when it comes to level design.
Thus we identify a problem of diversity in Procedural Level Generation via Machine Learning (PLGML). 

To address this problem, PLGML researchers have resorted to constructing their own training corpora.
For example, game level information can be represented as images \cite{9390320, chen2020image}, gameplay videos \cite{summerville2016learning}, or as abstractions of in-game object behaviour \cite{10.5555/3505288.3505295, DBLP:conf/aiide/SummervilleSSO20}. 
An example of this practice and a valuable contribution to the PLGML community is the Video Game Level Corpus (VGLC) \cite{summerville2016vglc}, which provides an annotated training corpora for level generation research. 
The VGLC represents a level with characters called tiles. 
A rich amount of literature has leveraged this representation to generate levels using various machine learning algorithms such as autoencoders, GANs, and LSTMs
\cite{DBLP:conf/digra/SummervilleM16,
sarkar2020exploring,
sarkar2020controllable, 
giacomello2018doom, thakkar2019autoencoder}. 
However, the significant amount of human effort that goes into converting game levels to this representation limits the number of games represented in this format.

Rather than relying on hand-authored representations for level design, recent research has looked into learning these representations \cite{karth2021neurosymbolic, mawhorter2021content, DBLP:journals/corr/abs-2206-06490, rabii2021revealing}. 
We previously introduced tile embeddings as a domain-independent vector representation of levels \cite{jadhav2021tile}. 
An autoencoder was trained to take in mechanical affordances and the local pixel context of a tile, and learned a representation unifying these pieces of information.
Tile embeddings have shown promising results in generating levels where the games have a good mix of tiles such as Lode Runner. 
However, we found that tile embeddings struggled to generate levels with imbalanced tile distributions.
For example, we observed that a tile embedding-based LSTM level generator for Super Mario Bros. resulted in empty levels (Figure \ref{fig:smb_level_generation}(b)).
This is a common problem in PCGML when the process of sampling new levels is greedy and biased towards the tile with the highest probability (in the case of SMB: empty sky tiles) \cite{snodgrass2013generating}. 

Traditional PLGML approaches have taken advantage of the discrete nature of the VGLC representation to alleviate the issue of skewed tile distributions. 
For instance, a level generator can be trained on the VGLC or any discrete representation such that given a sequence of previous tiles in a level, it predicts a distribution over the likelihood of possible next tiles. 
When generating a new level, tiles at each position can be sampled from this probability distribution \cite{DBLP:conf/digra/SummervilleM16}.
This sampling process solves the problem of producing empty levels encountered with a greedy tile selection strategy (Figure \ref{fig:smb_level_generation}(b)).

In order to enable sampling in our level generator, we learn a discrete representation by clustering learned tile embeddings.
Thus in the presented work, we leverage the benefits of learning simultaneous discrete and continuous representations to improve level generation for games with skewed tile distributions. 
This allows us to approximate the benefits of a discrete representation like the VGLC without the cost of hand-processing training data.
The main contributions presented in this work are as follows:
\begin{itemize}
    \item We introduce Cluster-based Tile Embeddings (CTE), which differ from our original tile embeddings \cite{jadhav2021tile} by the incorporation of edge information and a cluster-based loss.
    \item We present a novel two-step level generation pipeline based on discretizing our new embedding representation.
    \item We demonstrate and compare the performance of our CTE representation against both the original tile embeddings and the VGLC representation at the task of level generation for games with skewed tile distributions. 
    \item We demonstrate our approach's ability to generate levels for two games that no prior PLGML approach has attempted: Bugs Bunny Crazy Castle and Genghis Khan, based solely on images of their levels. 
\end{itemize}

\section{Related Work}
In this section we discuss prior work that has investigated the role of clustering in game level design, as our approach learns to discretize our tile embedding using clustering.    
Clustering is an unsupervised machine learning technique to discover groupings in data. 

\cite{guzdial2016learning} employed clustering to help learn probabilistic graphical models for Mario level design. 
\cite{DBLP:conf/aiide/Snodgrass18} proposed an approach to automatically identify sets of tiles, based on Markov Random fields and clustering. 
Similar to these approaches, we use clustering as part of our representation learning. However, these previous studies have based their clustering decisions solely on RGB representations. In our presented work, along with the RGB representation of a tile, we also incorporate behavioural and edge information.

\cite{yang2020game} employed a Variational Autoencoder with a Gaussian mixture as a prior distribution (GMVAE) for level generation.
Their work relies on clustering to identify similar $(16 \times 16)$ chunks from levels of multiple games. The learned components of the Gaussian Mixture Model are then used to generate new chunks of the same style. 
\cite{karth2021neurosymbolic} proposed neurosymbolic map generation using a VQ-VAE and Wave Function Collapse (WFC). 
A VQ-VAE quantizes patches of level images to a finite tileset on which WFC is applied to generate levels.
While \cite{karth2021neurosymbolic} focused on discretizing large patches of level design images, our work extracts representation of individual $16 \times 16$ tiles similar to \cite{yang2020game}
Like both approaches, our work also uses clustering for level generation. However, our approach differs by learning continuous and discrete representations of tiles and utilising both  for level generation.

To the best of our knowledge, we are the first to tie clustering and embeddings together for representation learning in PCGML. However, this approach has been explored in other fields like reinforcement learning for games. \cite{NEURIPS2020_d90e5b66} introduced the shrinkage effect in training an encoder for extracting representations of players in professional ice hockey. 
It allows the model to transfer information between the observations of different players such that statistically similar players lead to similar representations under similar game contexts. 
We draw a parallel to this work and implement clustering loss to enforce intrinsic clustering and assign similar representations to tiles with similar RGB pixel representation, affordances and edges.

\section{System Overview}
The goal of this work is to learn an improved tile embedding for games with skewed tile distributions for the task of level generation. 
Towards this objective, we begin this section by discussing our modifications to the original tile embedding autoencoder to learn our new Cluster-based Tile Embeddings (CTE). 
Next, we explain the limitations of an LSTM level generator trained on the original tile embedding representation for games with skewed tile distributions. 
We then present our novel two-step level generation pipeline that learns a discretization of our CTE through clustering and leverages both representations for level generation.

\begin{figure*}[tbh]
    \centering
     \includegraphics[scale=0.24]{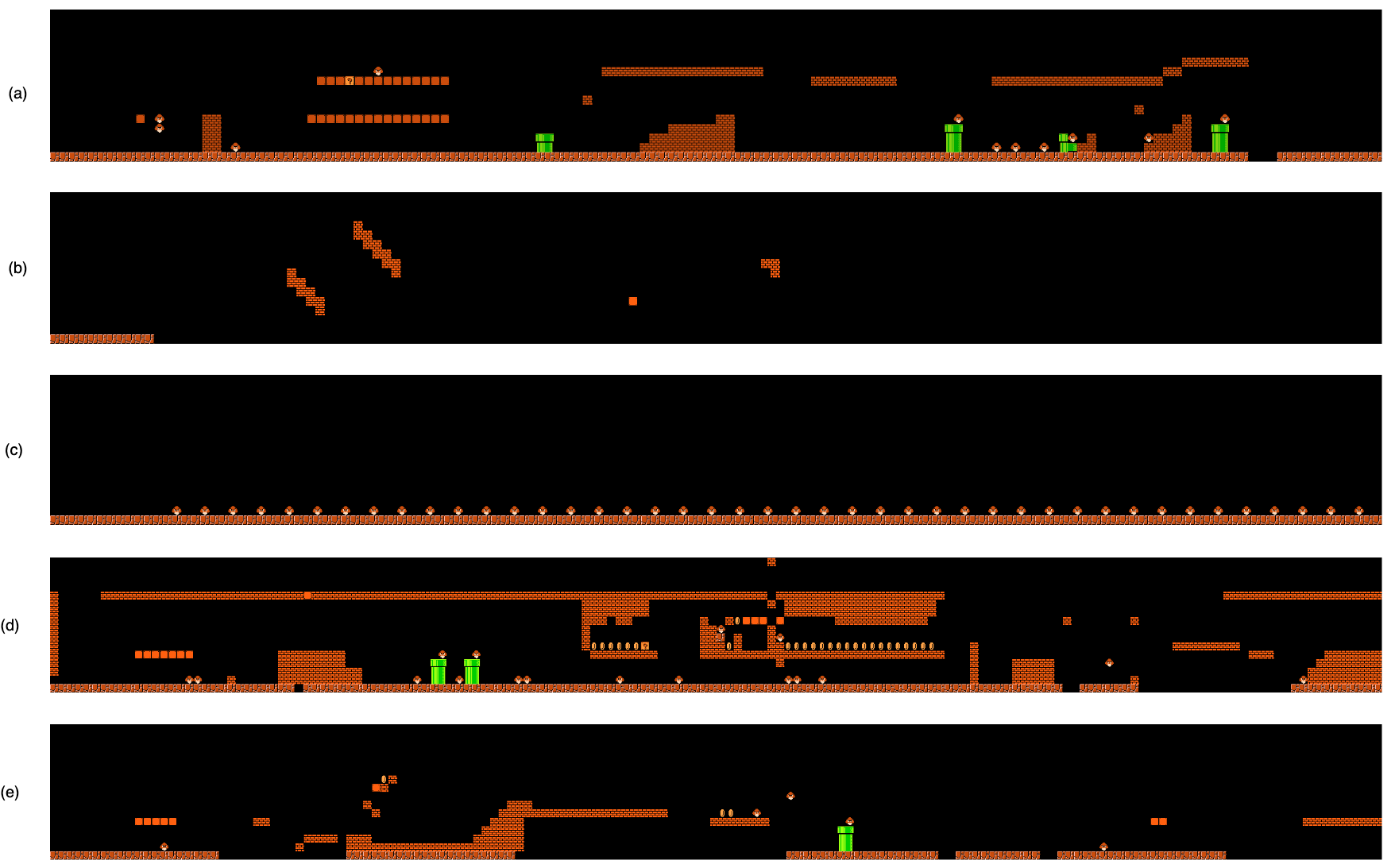}
     \caption{SMB LSTM level generator outputs with: (a) VGLC representation (b) original tile embedding (c) CTE. We also include good (d) and bad (e) examples for our two-step CTE level generation process.}
     \label{fig:smb_level_generation}
\end{figure*}

\begin{figure*}[tbh]
    \centering
     \includegraphics[scale=0.33]{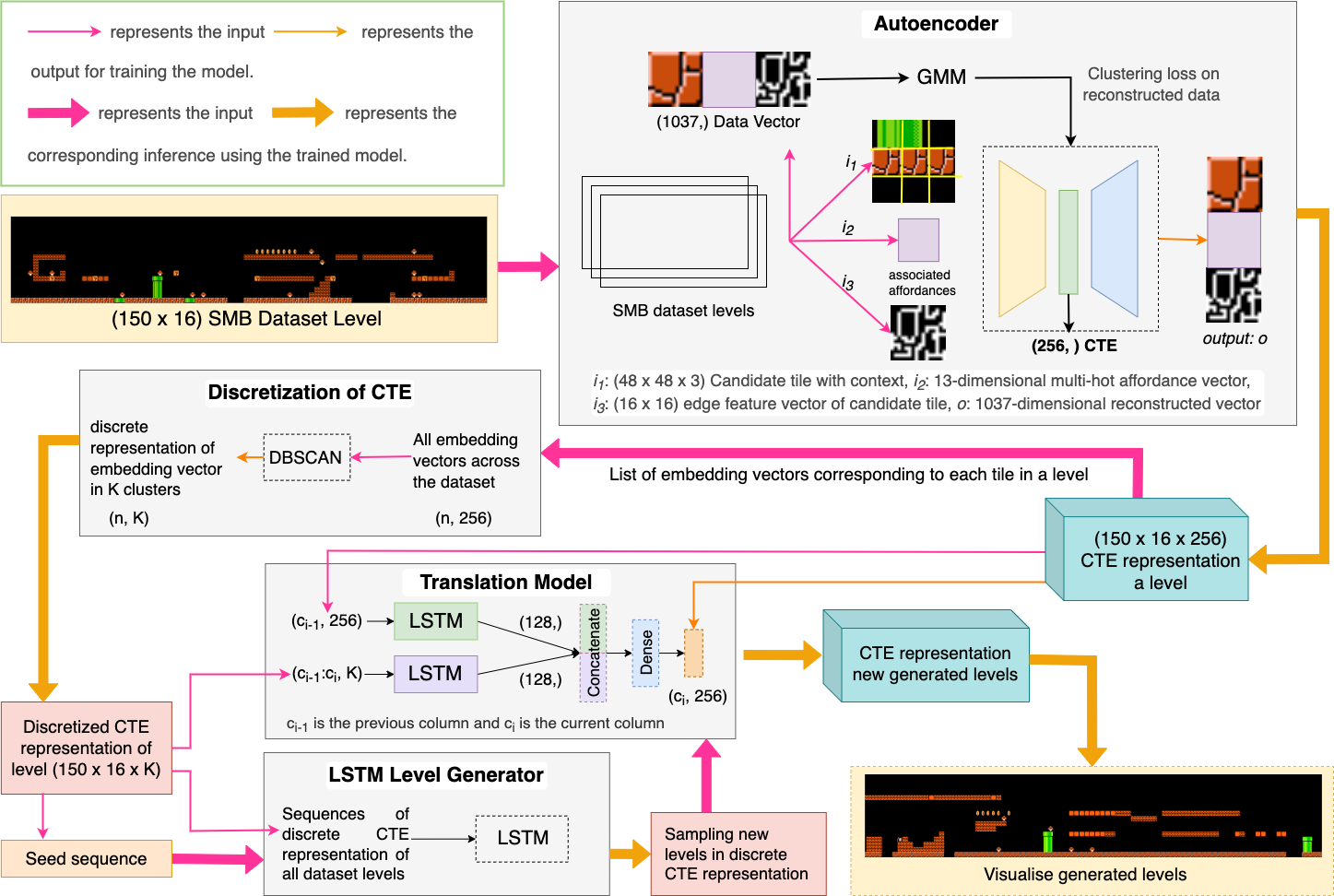}
     \caption{A complete system diagram. We train an autoencoder on the RGB, affordance, and edge information using a cluster-based loss to learn our Cluster-based Tile Embedding (CTE). We then discretize this representation via DBSCAN, and train an LSTM level generator on this discretized CTE. We train a translation model (also an LSTM) to convert back to CTE from the discrete representation output by the level generator.}
     \label{fig:system_diagram}
\end{figure*}

\subsection{CTE: Cluster-based Tile Embeddings}
The VGLC tile-based representation of a level $L$ is an $h \times w$ dimensional array. Here $h$ and $w$ are the height and width of the level, respectively.
Each character of $L$ is called a tile which is associated with a $16 \times 16$ pixel representation in a level image and a corresponding set of affordances. Affordances convey a tile's mechanical behaviour.

Our original tile embedding work employed a dual branched autoencoder to learn a 256-dimensional embedding vector representation of a tile \cite{jadhav2021tile}. The network accepted two inputs: 1) a 3*3 grid of the candidate tile at the centre with its neighbours surrounding it in the 16*16*3 RGB pixel representation $(48\times48\times3)$, 2) the candidate tile's 13-dimensional one-hot affordance vector.
To compare more easily to the original tile embedding work, we utilise the same set of games (Super Mario Bros., Kid Icarus, Megaman, Lode Runner and Legend of Zelda) as our training corpus and maintain the same tile-affordance mapping. 
The tile-based level data is taken from the VGLC corpus\footnote{\url{https://github.com/TheVGLC/TheVGLC}} and the JSON files for tile-affordance mapping are from the original tile embedding implementation\footnote{\url{https://github.com/js-mrunal/tile\_embeddings}}. 
We make two modifications to the training of the original autoencoder to better handle level design tasks for games with skewed tile distributions and refer to the newly extracted 256-dimensional embedding vector as the \textbf{Cluster-based Tile Embedding} (CTE).

\subsubsection{Incorporating Edge Information:} When applying the original tile embeddings to games where the affordance information was unknown, we found that the latent space representations depended predominantly on coloured pixel information of a tile.
For instance, an empty blue sky tile was placed close to a solid blue brick tile.
To discourage this, we included edge information into our embedding.
Canny edge detection \cite{canny1986computational} is a common algorithm for identifying edge information. 
We convert the $16 \times 16 \times 3$ pixel representation of a tile to grayscale and apply the canny edge detection algorithm to obtain a $16 \times 16$ edge feature vector. 
Thus for each candidate tile, we feed three inputs to our autoencoder: the pixel representation of the candidate tile along with its neighbours $(48\times48\times3)$, a 13-dimensional multi-hot affordance vector and $(16\times16)$ edge features.

\subsubsection{Clustering Loss: }In the original tile embedding work, the learned latent space was fairly continuous, without clear separation between types of tiles.
Learning more distinct groups can improve the utility of a final representation \cite{7471631}. 
With an aim to push representations of similar elements closer while keeping representations of dissimilar elements apart, we introduce an explicit cluster-based loss $L_c$ in the training process.
For this cluster-based loss, we must cluster our data prior to training our autoencoder. The idea is to leverage the clusters as a guide for representation learning.
For each candidate tile, its $16 \times 16 \times 3$ RGB pixel representation, 13-dimensional multi-hot affordance vector, and $16 \times 16$ edge vector are fed to a Gaussian Mixture Model (GMM) \cite{reynolds2009gaussian}. 

A tile can belong to multiple clusters. For instance, it is appropriate to assign a Cannon \includegraphics[scale=0.6]{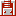} in MegaMan to a cluster of \textit{Hazards} as well as to a cluster of \textit{Solids}. We rely on a GMM in order to account for such potential overlap in tile groups. We pick an elbow point based on the Silhouette score and Bayesian Information Criterion (BIC) to determine the optimal number of clusters \cite{10.1016/0377-0427(87)90125-7, 10.1214/aos/1176344136}. 
For the given VGLC dataset, we observe an elbow point at 10 clusters.

We compute our clustering loss ($L_c$) as the categorical cross entropy error between the GMM cluster assignment of a given tile and its corresponding embedding during training. Along with $L_c$, our loss function includes the mean squared error on the reconstructed edge feature vector ($L_e$), the mean squared error over the reconstructed image data ($L_i$) and the binary cross entropy loss on the reconstructed affordances ($L_a$).
In totality, the loss function can be mathematically represented as:
\begin{equation}
Total\ loss\ =\ (0.5\ L_i)+(1.5\ L_a)+(0.5\ L_e)+(0.5\ L_c)
\end{equation}
To accurately embed affordance information, we increase the relative weight of its reconstruction.

\subsection{Level Generation for Super Mario Bros.}
In this section, we describe the difficulty in generating SMB levels using an LSTM trained on the original tile embeddings and CTE, which motivated our novel two-step level generation process described below. 
\subsubsection{Problems with SMB Level Generation: } 
We train two LSTM models, one on the original tile embeddings and the other on our CTE representation, for SMB. We follow the training process from \cite{jadhav2021tile}. 
Sampling from an LSTM trained on a continuous representation is deterministic and hence for a given seed input, these models generate only one output as shown in Figure \ref{fig:smb_level_generation}(b) and (c) respectively. 
In both cases we feed in the same 200 tiles of flat ground as input. 
While the CTE representation helps the LSTM learn to generate more reasonable output than the original tile embedding, the output is repetitive and does not reflect Mario-like structure.
These outputs show clear limitations of an embedding-based generator in modelling levels with skewed tile distributions, given that the outputs for games with balanced tile distribution appear much more like the original levels \cite{jadhav2021tile}. 
In Table \ref{table:skewed_balanced} of the Appendix, we outline the difference in tile distributions for a skewed and a comparatively balanced tile-based game.

\textbf{Analysis: } A possible explanation for these poor results is the lack of a sampling mechanism in the generator, since the CTE output is similar to the most likely Mario level output from a probabilistic generator \cite{snodgrass2013generating}. 
In comparison, we observe higher quality results when our LSTM is trained on the VGLC representation as seen in Figure \ref{fig:smb_level_generation}(a). 
The only difference between the two models is in the output layer \cite{jadhav2021tile}.
For the model trained on the VGLC representation, the output layer is a probability distribution $p$ over possible tile types \cite{DBLP:conf/digra/SummervilleM16}. The next tile is sampled from $p$. If we simply pick the most likely next tile, we output levels similar to Figure \ref{fig:smb_level_generation}(c) even with the VGLC representation.
We cannot sample from an LSTM trained on either tile embedding, as the LSTM would output the closest tile embedding, not a probability distribution. 
To remedy this, we present a two-step level generator which discretizes CTE.

The two steps of this level generator are to first generate levels in a discrete representation, allowing sampling to occur. 
Then we have a secondary translation model that converts the levels in this discrete representation back into our CTE representation, so that we can visualize them and extract the predicted affordances. 
This two-step level generation process naturally requires training two distinct models, one for each step. 
For both models we use the same LSTM architecture used throughout this paper.

\subsubsection{Step I. Training LSTM on Cluster Levels: }
To obtain the discrete representation of each level, we leverage the latent structure imposed by the clustering-based loss function. 
We first begin by converting each level to the CTE representation and then feed all the CTE embedding vectors to the density-based clustering algorithm, DBSCAN \cite{10.5555/3001460.3001507}.
Unlike partitioning-based and distribution-based clustering algorithms, DBSCAN has a number of benefits for clustering in a latent space \cite{DBLP:conf/secrypt/KellerM0Y21}, which makes it highly relevant to this task. 

Figure \ref{fig:system_diagram} shows the overview of our system architecture. If we consider an SMB level of $150\times16$ VGLC tiles, and replace each tile with its 256-dimensional embedding, we get the $150\times16\times256$ dimensional CTE representation. 
Further, if each of these embeddings is assigned to a cluster we can simply represent a 256-dimensional embedding by a cluster identifier to get a $150*16*K$ discretized representation, where $K$ is the optimal number of clusters.
We refer to this as our cluster representation. For SMB, the optimal number of clusters (K) detected by DBSCAN was 11 with a silhouette score of 0.91. 
We note that we recalculate these clusters for each new game, unlike the clusters used to inform the CTE cluster loss. 
We note that we do this to learn a discrete representation as we cannot use the VGLC level representation directly or generating levels for games outside the VGLC corpus would be impossible. 

\subsubsection{Step II. Translation Model: }
The generated output of the previous step is in the cluster representation and cannot be used directly.
A cluster may consist of many member tiles, thus a cluster identifier may not be adequate for accurate visual and affordance reconstruction. 
Therefore, we need a translation mechanism to convert the cluster representation of a level to its associated CTE representation.
We train an LSTM network to translate from the cluster representation to CTE.
Such a translation mechanism requires the knowledge of context as well as affordances. For instance, to rebuild a solid red brick tile pattern, red bricks cannot be followed by blue bricks even though they may belong to the same cluster.
Therefore, as illustrated in Figure \ref{fig:system_diagram}, a CTE representation of column $c$, depends on: a) the underlying cluster representation of column $c$ and $c-1$, b) the CTE representation of column $c-1$. 
With this approach, we observed instances where the translation model did not output CTE tiles from the correct clusters.
Thus, we replace translated CTE output with its nearest neighbour from the correct cluster. Translated SMB test dataset levels are shown in the Appendix.

\section{Experiments}
We evaluate our two-step level generation pipeline and CTE representation by sampling levels for Super Mario Bros., a game with a skewed tile distribution. We employ commonly-used PCGML metrics to assess the quality of our outputs in comparison to the outputs of an LSTM trained on the VGLC representation, original tile embeddings, and CTE. Additionally, we test our approach's ability to represent and generate levels for two unannotated games: Bugs Bunny Crazy Castle and Genghis Khan. In this section, we describe our experiments and report our results.

\begin{table*}[!tbh]
\centering
\small
\begin{tabular} {cc|cccc}
\hline
& \textbf{Dataset} & \textbf{LSTM on VGLC} & \textbf{Two-step level generation} & \textbf{LSTM on CTE} & \textbf{LSTM on original} \\
&&&&&\textbf{tile embeddings}\\
\hline
\hline
Leniency & -0.0069 $\pm$ 0.0084 & -0.0096 $\pm$ 0.0077 & \textbf{-0.0054 $\pm$ 0.0102} & -0.0021 $\pm$ 0.0078 & \textit{-0.0130 $\pm$ 0.0155} \\
\hline
Density & 0.1315 $\pm$ 0.0642 & 0.1669 $\pm$ 0.0654 & \textbf{0.1625 $\pm$ 0.0600} & 0.0485 $\pm$ 0.0310 & \textit{0.0721 $\pm$ 0.0172} \\
\hline
Linearity & 0.0515 $\pm$ 0.0729 & 0.0362 $\pm$ 0.0514 & \textbf{0.0466 $\pm$ 0.0737} & 0.7234 $\pm$ 0.3540 & \textit{0.8208 $\pm$ 0.3435} \\
\hline
Interestingness & 0.0254 $\pm$ 0.0133 & \textbf{0.0279 $\pm$ 0.0114} & 0.0227 $\pm$ 0.0082 & 0.0002 $\pm$ 0.0005 & \textit{0.0002 $\pm$ 0.0003} \\
\hline
Enemy Sparsity & 42.0036 $\pm$ 17.512 & \textbf{34.6699 $\pm$ 7.7747} & 32.3738 $\pm$ 10.4389 & 0.25 $\pm$ 0.25 & \textit{0.0 $\pm$ 0.0} \\
\hline
Playability & 86.4864  & 54.0 & 40.0 & \textbf{100.0*} &  0.0*  \\
\hline
\end{tabular}
\caption{Comparative study of SMB generators based on PCGML tile metrics. \textbf{Bold} indicates mean values nearest and \textit{Italic} indicates values farthest from the original mean dataset values. Asterisks indicate theoretical values.}
\label{table:tile_metrics}
\end{table*}

\subsection{Level Generation for SMB}
The training corpus for this experiment consists of the 37 levels from Super Mario Bros. and Super Mario Bros. 2 (Japan) from the VGLC Corpus \cite{summerville2016vglc}.
We analyze the performance of our two-step level generator for SMB level generation and compare it against the results of LSTMs trained directly on the original tile embeddings, CTE, and the VGLC representation \cite{jadhav2021tile, DBLP:conf/digra/SummervilleM16}.
For all the level generation models the history sequence is maintained at 200 tiles and the network consists of three layers each comprised of 512 LSTM cells. 
We partition the data as 80-10-10\% train, test and validation split. 
LSTMs trained directly on the original tile embeddings and CTE output the continuous embedding vector of the next tile whereas the LSTMs trained on discrete CTE (two-step level generation) and the VGLC output a distribution over tiles with softmax activation at the final layer. This makes sampling possible. The new levels are sampled tile by tile by generating rows left to right then bottom to top.

Ideal output levels would match the style of existing SMB levels. \cite{10.1145/1814256.1814260} suggested several metrics to assess the style of generated content in comparison to the dataset. 
\begin{itemize}
    \item \textbf{Leniency }captures the difficulty of the level. Values closer to one indicate more lenient levels \cite{10.1145/1814256.1814260}. 
We compute leniency as,
\begin{equation}
    leniency\ = \frac{2r-(0.5*g)-e}{T}
\end{equation}
where $r$, $g$, and $e$ represent the counts for rewards, gaps, and enemies respectively, and $T$ is the total number of tiles in a level $(l \times w)$. 

\item \textbf{Linearity }measures how well a level fits to a line. It is calculated as the mean squared error between the centre points of each platform and its projection on the linear regression line \cite{10.1145/1814256.1814260}. 

\item\textbf{Interestingness }is an important metric especially for evaluating generators for skewed tile distributions because the most probabilistic tile is unlikely to be interesting. It measures the fraction of tiles that bring visual variety to the level \cite{10.1145/3102071.3102080}. 

\item \textbf{Density }
is the proportion of solid tiles in the level. Density is a relevant here, as we observe that it is possible for SMB generators to produce only empty tiles because of their high probability  \cite{10.1145/3102071.3102080}.

\item \textbf{Enemy Sparsity } measures the horizontal spread of the enemies across the level \cite{10.1145/3102071.3102080}.
Because SMB levels include lines of enemies, it is possible for a generator to get stuck generating a continuous string of enemies. 
We calculate enemy sparsity as:
\begin{equation}
    Enemy Sparsity= \frac{\sum_{e \in E} |x(e)-\overline{x}|}{|E|}
\end{equation}

where $x$ is the $x$-position of an enemy, $\overline{x}$ the average $x$-position of enemies, and $|E|$ the total number of enemies. 

\item \textbf{Playability} measures the percentage of playable levels generated. We run an A* agent provided in the VGLC to check for the existence of path in a level \cite{summerville2016vglc}. 
\end{itemize}

As illustrated in Figure \ref{fig:smb_level_generation}, we observe a notable improvement in the quality of levels generated by our proposed two-step level generator with CTE in comparison to the LSTM on original tile embeddings and LSTM on CTE. 
Compared to the good examples of level generation, the bad ones are fairly empty and consist of unreachable sections because of large platform gaps or height (Figure \ref{fig:smb_level_generation}(e)). Meanwhile, the good examples show the presence of more interesting tiles and have a coherent structure better matching the style of the dataset (Figure \ref{fig:smb_level_generation}(d)). But these are only two examples.

\begin{figure*}[tbh]
    \centering
     \includegraphics[width=\textwidth]{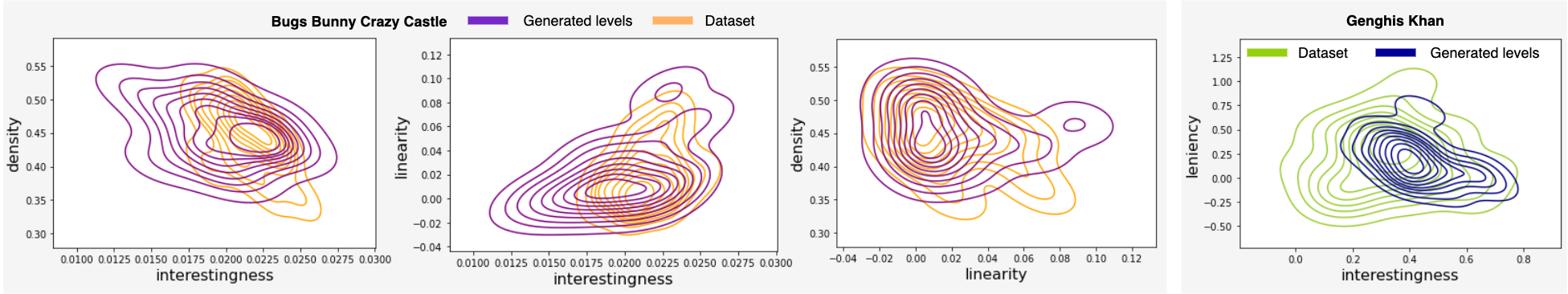}
     \caption{Expressive range analysis for the unseen games.}
     \label{fig:unseen_games_kde}
\end{figure*}

Table \ref{table:tile_metrics} shows the results of the metrics-based evaluation between 50 output levels generated by our two-step level generator, LSTM on VGLC representation, LSTM on original tile embeddings, LSTM on CTE and the original SMB dataset. 
Level generation using discrete representations (VGLC and discrete-CTE) consistently outperforms level generation using continuous representations (original tile embedding and CTE) across all tile metrics. 
The distribution of levels generated by the LSTM trained on the original tile embeddings and the LSTM trained on CTE is nowhere close to the distribution of the original dataset. This is also evidenced in Figure \ref{fig:smb_level_generation} (b) and (c).
We take this to indicate that the two-step generation process allows CTE to compete with the hand-authored VGLC representation.

These results reinforces the importance of a discrete representation and sampling in level generation for levels with skewed tile distributions. 
Further, our two-step level generator's levels more closely resemble the training distribution compared to the VGLC generator levels in Leniency, Density and Linearity. 
In a similar vein, although the VGLC generator outperforms our approach for the other two metrics, we find the performance comparable. 
The playability results are an oddity, since there are levels the provided A* agent cannot complete in the original dataset but the LSTM on CTE levels (because they only include flat ground) can always be completed. 
On the other hand, the VGLC and discrete-CTE level playability values are comparatively close.
We find these results valuable as unlike discrete-CTE, the VGLC benefits from being human-authored.

Approximating the actual distribution of game levels accurately is difficult given the limited size of the test split. Therefore the Dataset column summarizes metrics across the entire dataset. 
To provide evidence that the model is not overfit, we report the minimum tile edit distance between cluster representations of generated levels and the dataset levels of games in the Appendix.

\subsection{Level Generation for Unseen Games}
We train our two-step CTE level generator to generate levels for two unseen games: Bugs Bunny Crazy Castle (BBCC) and Genghis Khan. 
We downloaded 20 levels of BBCC and 41 of Genghis Khan as our training corpus\footnote{\url{https://vgmaps.com/}}.
We chose these particular games because of their contrasting degree of structure variance, with BBCC being comparatively higher. 
For both games the affordance information is missing. In such cases, the clustering relies on visual and edge data.

BBCC is an action-puzzle NES game where the player moves Bugs through rooms collecting carrots. It has a set of representative tiles consisting of solid brick patterned background \includegraphics[scale=0.6]{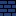}; boxing gloves \includegraphics[scale=0.6]{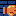}, invincibility potions \includegraphics[scale=0.6]{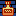}, safes \includegraphics[scale=0.6]{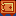}, crates \includegraphics[scale=0.6]{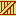}, and ten thousand-pound weights \includegraphics[scale=0.6]{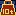} that can be used against the enemies in the game; and solid tiles \includegraphics[scale=0.6]{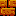}, \includegraphics[scale=0.6]{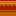}, \includegraphics[scale=0.6]{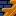}, \includegraphics[scale=0.6]{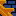} on which bugs can stand. 
Genghis Khan is a turn-based strategy game. Its tiles exhibit comparative similarity in structure as well as colour. 
Thus, generating levels for both games allows us to study the impact of structural variance in our learned representation.

We train a two-step level generator on both games by employing a similar process as for SMB. 
The only difference is that we pass a zero vector for the affordances when extracting the CTE representation. 
We found the optimal number of clusters for the two-step generator was 8 and 24 for BBCC and Genghis Khan respectively. Examples of output levels are given in the Appendix.

To evaluate the performance of the generator, we employed expressive range analysis on the generated content in comparison to their respective datasets \cite{10.1145/1814256.1814260}. 
Due to the limited size of test split, we use the entire dataset to estimate the true distribution.
We perform expressive range analysis on Interestingness, Linearity and Density for BBCC. For Genghis Khan, we approximate only Interestingness and Leniency. We do not calculate Linearity and Density since it is not a platformer game. 
The BBCC metrics remain unchanged, as the game is sufficiently similar to SMB.
Interestingness in BBCC is calculated as the frequency of tiles representing the items. Similarly, interestingness in Genghis Khan is measured as the proportion of tiles that bring visual variety to levels such as mountains \includegraphics[scale=0.3]{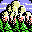}, forests \includegraphics[scale=0.3]{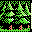}, towns \includegraphics[scale=0.3]{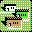}, and castles \includegraphics[scale=0.3]{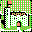}. Further, we use the movement cost associated with each tile to calculate the Leniency. We assign negative costs to tiles that are difficult to move across such as -5 for mountains and castles, -6 for deserts, and -8 for rivers. In comparison, it costs less to move across regular land, forests and towns thus we assign movement costs of 3, 3, and 4 respectively. We calculate Leniency by summing these movement costs normalized by the total number of tiles in a level.

Figure \ref{fig:unseen_games_kde} shows the expressive range analysis performed on the generated levels of unannotated games in comparison to the entire original datasets.
For BBCC, our model covers a considerable amount of the generative space, with a large amount of overlap with the original levels. 
However, we found lower output interestingness than the true distribution.
For BBCC, the density of levels increases as linearity decreases, this is due to the fact that as more platforms are generated vertically, levels become denser due to the presence of platforms and stair tiles.

In comparison to BBCC, the Interesting-Leniency expressivity analysis for Genghis Khan does not match the training distribution as closely, though there is still significant overlap. We find that the generated outputs have more challenge, more difficult terrain, compared to the training dataset. 
Although these results can be improved, we find them promising, indicating the capability of the generator to design levels for games based only on image data.

\section{Discussion, Limitation and Future Work}
The modified Cluster-based Tile Embeddings (CTE) and the two step level generation pipeline demonstrated improved performance in generating levels for Super Mario Bros., and the ability to generate levels for unseen games.
Notably, our approach also shows potential in generating levels of non-platformer games such as Genghis Khan, a turn-based strategy game. 
However, we can still improve our pipeline further, especially for games with structurally similar tiles and missing affordances.

We employ Silhouette Score and the Structural Similarity Index to evaluate the performance of the Clustering and Translator modules in the two-step level generation pipeline. 
The results of these evaluations are in Appendix. We find that clustering is a crucial component of our level generation pipeline and the performance of the translation model improves with the quality of the clustering.
For Genghis Khan, the missing affordances and lack of structural variability between representative tiles yielded a low silhouette score, thus implying arbitrary clustering. 
If the cluster participants have no particular structure, we observe that the translation model struggles to map cluster identifiers to embeddings.

We propose the following two avenues to explore in terms of architecture to learn stronger representations: 
(1) In the presented work, we approximate the mixture of Gaussians before training an autoencoder. The recent trend in structuring the latent space has been towards Deep Clustering i.e., learning the embedding representations that optimize clustering and performing cluster assignment simultaneously. In a similar vein, a GMVAE imposes a mixture of Gaussians as a prior on the latent space, which could be used to learn a more robust representation \cite{yang2020game}. 
(2) Discrete representations have benefits over continuous representation for several PCGML tasks. The VQ-VAE is a variant of the VAE that quantizes the latent space to learn a discrete latent representation \cite{DBLP:journals/corr/abs-2203-12130}. Leveraging a VQ-VAE could potentially simplify our current level generation pipeline. Such an implementation opens the possibility to having a common discrete representation across multiple games and thus learning a generalized level generator. 

Before applying a VQ-VAE to learning tile embeddings, we would need to address a number of caveats. In our current work, we reflect on the idea that continuous and discrete representations are both needed for level generation. While discrete representations are a natural fit for many applications, learning only discrete representations can limit the expressivity of the generator. Further, these representations cannot be applied directly in tasks based on interpolating between points in a learned latent space. For example, in generating novel tiles. This might be relevant in another future application of CTE: the PLGML task of level blending \cite{sarkar2021generating}.

\section{Conclusions}
In this work, we presented a novel architecture and training process to learn new Cluster-based Tile Embeddings (CTE). To improve the quality of the embeddings, especially for unseen games, we incorporate edge features and a new cluster-based loss. While tile embeddings perform poorly at generating levels for games with skewed tile distributions, we propose a two-step level generation process to address this problem. We demonstrate a strong performance at generating Super Mario Bros. levels as well as two unseen games: Genghis Khan and Bugs Bunny Crazy Castle.

\section*{Acknowledgments}
We acknowledge the support of the Natural Sciences and Engineering Research Council of Canada (NSERC) and Alberta Machine
Intelligence Institute (Amii).

\bibliography{References}

\appendix
\begin{table*}[ht]
\centering
\small
\begin{tabular} {ccc}
\hline
\textbf{Game} & \textbf{Train} & \textbf{Test} \\
\hline
\textbf{SMB} & 768.92 $\pm$ 140.98 & 743.06 $\pm$ 101.67 \\
\hline
\textbf{Genghis Khan} & 39.32 $\pm$ 3.93 & 40.98 $\pm$ 5.18 \\
\hline
\textbf{BBCC} & 192.39 $\pm$ 22.30 & 201.84 $\pm$ 19.24 \\
\hline
\end{tabular}
\caption{The edit distances observed between the generated cluster representations and the training and test data suggests that the model is not overfitting.}
\label{table:edit_distance}
\end{table*}

\begin{table*}[ht]
\centering
\small
\begin{tabular} {cccc}
\hline
 & 
\multicolumn{2}{c}{Clustering} & Translation \\[1.0ex]
& Number of Clusters & Silhouette Score &  Structural Similarity Index\\
\hline
\hline
SMB & 11 & 0.91 & 0.9976 $\pm$ 0.0014\\
\hline
Genghis Khan & 8 & 0.39 & 0.8689 $\pm$ 0.01\\
\hline
BBCC & 24 & 0.53 & 0.9792 $\pm$ 0.0079\\
\hline
\end{tabular}
\caption{Evaluating Clustering and Translator Modules using Silhouette Score and SSIM.}
\label{table:eval_clus_translation}
\end{table*}
\begin{figure*}[ht]
    \centering
     \includegraphics[width=4in]{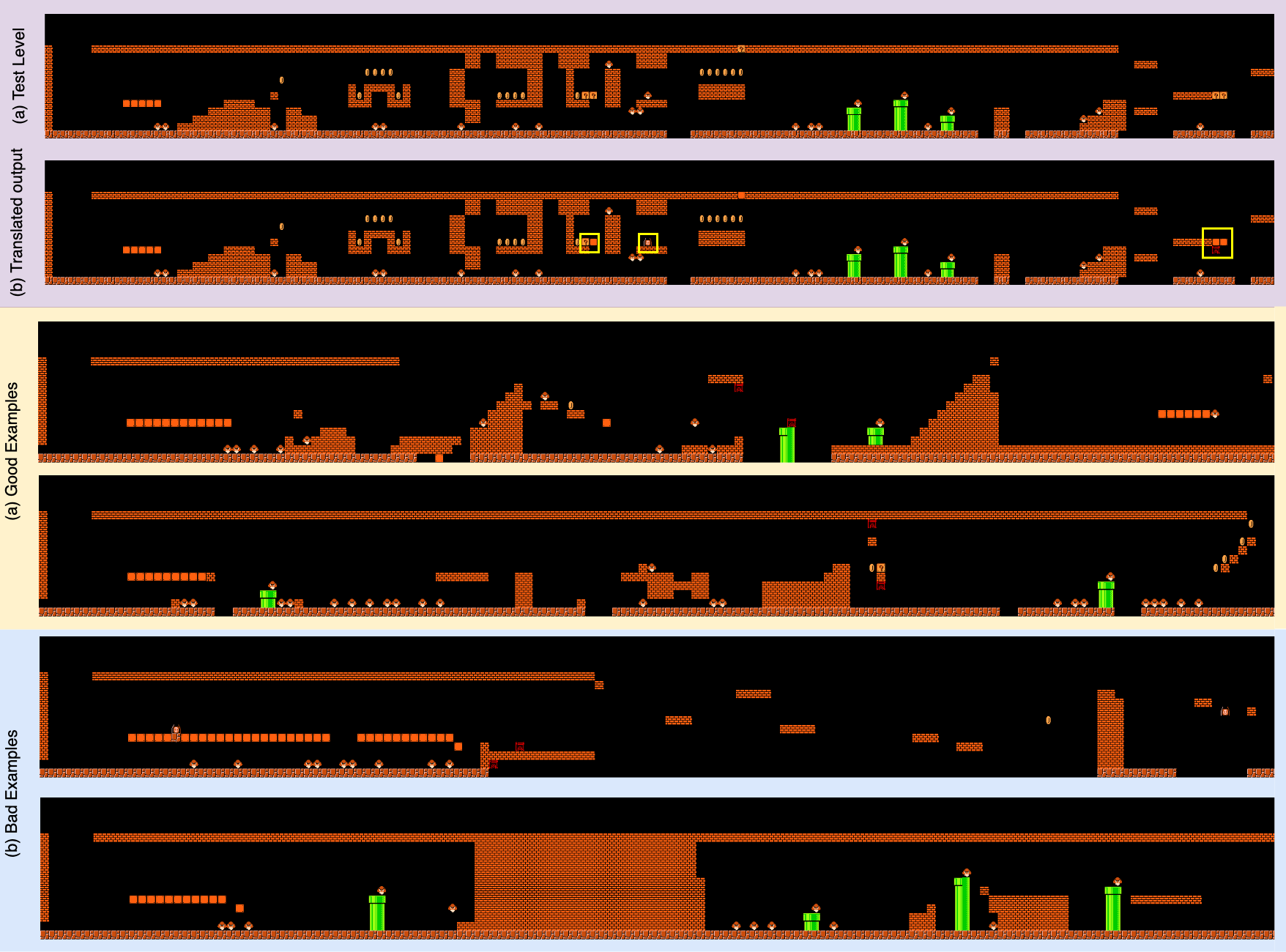}
     \caption{Figure (a) shows a test SMB dataset level and Figure (b) shows its translated output obtained using the second step of our two-step generator. The differences between the two are highlighted in yellow. To get this translated version we convert the dataset levels of a game to: 1) their cluster representation using the DBSCAN and 2) their CTE representation using our newly trained autoencoder. We use these cluster representation and their correspoding CTE representation of dataset levels to train the translation model as discussed in the two-step level generation process. 
     Figure (c) and (d) show more examples of SMB level generation output with the two-step level generator trained on our CTE representation.}
     \label{fig:more_smb_outputs}
\end{figure*}

\begin{figure*}[tbh]
    \centering
     \includegraphics[width=6.5in]{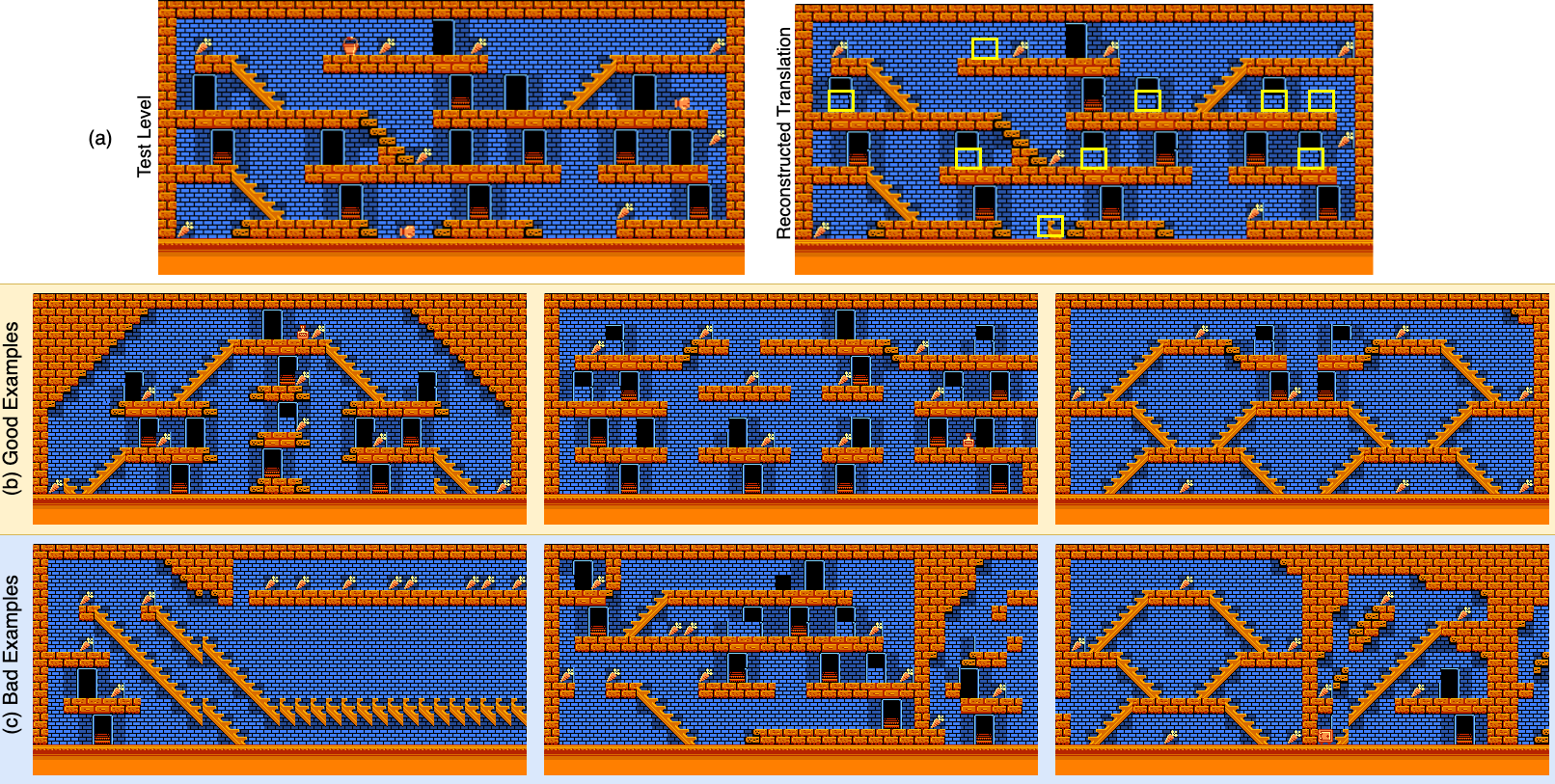}
     \caption{Level Generation for Bugs Bunny Crazy Castle : (a) Test dataset level (left) and its corresponding translated output (right) with differences highlighted in yellow (b) Examples of good generation output (c) Examples of bad generation output. Unlike good examples as in shown in (b), bad examples in (c) show the presence of unreachable level sections due to the lack of portals/doors, and inconsistency in level structure.}
     \label{fig:bbcc_outputs}
\end{figure*}

\begin{figure*}[tbh]
    \centering
     \includegraphics[width=5.5in]{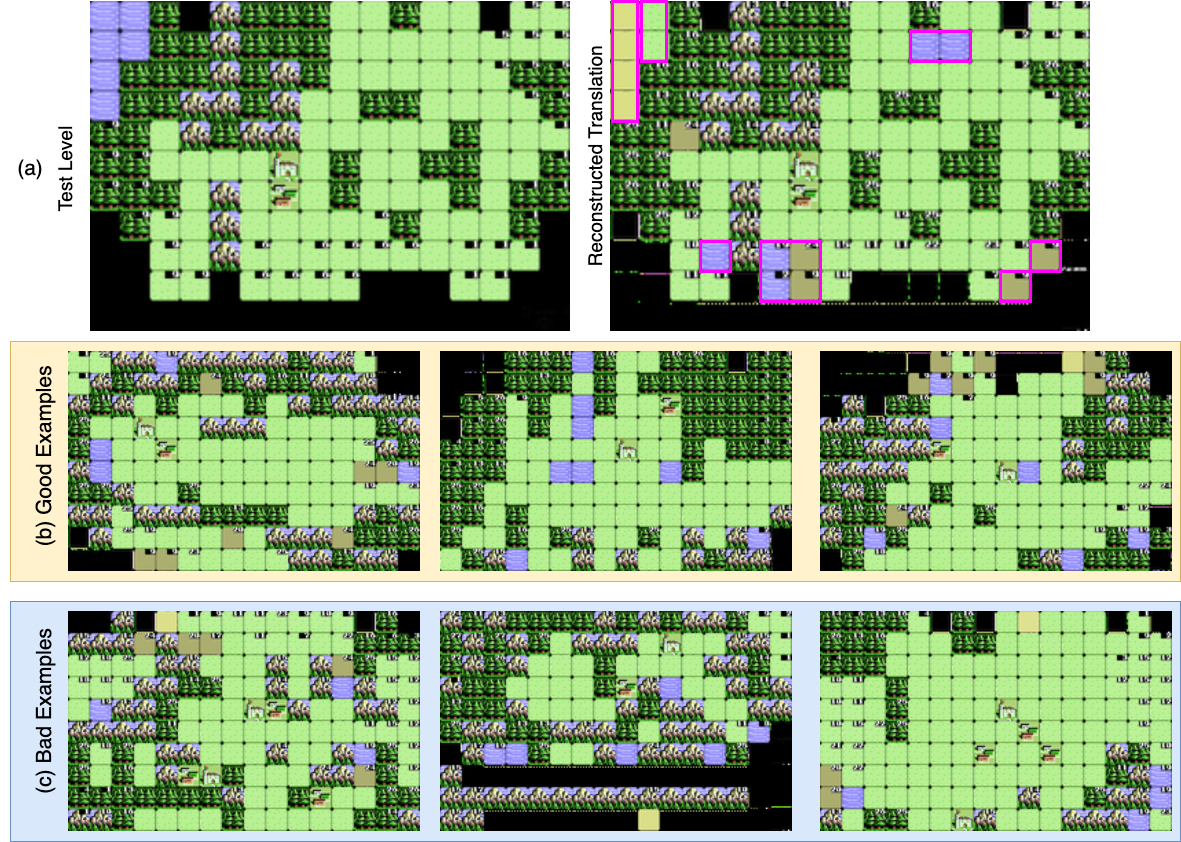}
     \caption{Level Generation for Genghis Khan : (a) Test dataset level (left) and its corresponding translated output (right) with differences highlighted in red (b) Example of good generation output(c) Example of bad generation output. The dataset levels of Genghis Khan only have one pair of town and castle tiles each whereas examples of bad generation (c), have multiple pairs. The bad levels also contain randomly placed mountain and forest tiles, instead of the clustered appearance found in (b) and in the original dataset.}
     \label{fig:gk_outputs}
\end{figure*}

\begin{table*}[!tbh]
\centering
\small
\begin{tabular} {ccc|ccc}
\hline
\multicolumn{3}{c|}{Super Mario Bros} & \multicolumn{3}{c}{Lode Runner} \\[1.0ex]
Tile & Example Tile Sprite & Median &  Tile & Example Tile Sprite & Median \\
\hline
\hline
-& \includegraphics[scale = 0.6]{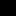}, \includegraphics[scale = 0.6]{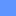} & 88.33\% & . & \includegraphics[scale=0.6]{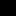} & 58.09\% \\
\hline
E & \includegraphics[scale = 0.6]{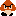}, \includegraphics[scale = 0.6]{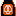}, \includegraphics[scale = 0.6]{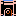} & 7.26\% & E & \includegraphics[scale=0.6]{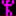} & 21.59\% \\
\hline
S & \includegraphics[scale = 0.6]{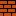} & 0.99\% & G & \includegraphics[scale=0.6]{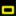} & 8.52\% \\
\hline
X & \includegraphics[scale = 0.6]{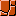} & 0.55\% & b & \includegraphics[scale=0.6]{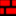} & 4.11\% \\
\hline
$<$ & \includegraphics[scale = 0.6]{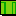} & 0.51\% & $\#$ & \includegraphics[scale=0.6]{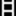} & 3.26\% \\
\hline
\end{tabular}
\caption{Median percentages of top five tiles occurring in a level. This table illustrates skewed tile distribution in Super Mario Bros and comparatively balanced tile distribution in lode runner tiles.}
\label{table:skewed_balanced}
\end{table*}

\end{document}